%% file: acl_latex.tex
\documentclass[11pt]{article}

\usepackage[final]{acl}

\usepackage{times}
\usepackage{latexsym}

\usepackage[T1]{fontenc}

\usepackage[utf8]{inputenc}

\usepackage{microtype}

\usepackage{inconsolata}

\usepackage{graphicx}
\usepackage[table]{xcolor}
\usepackage{microtype}
\usepackage{graphicx}
\usepackage{caption}
\usepackage{subcaption}
\usepackage{booktabs} 
\usepackage{hyperref}
\usepackage{times}

\usepackage{lineno}
\usepackage{amsmath}
\usepackage{amsfonts}
\usepackage{algorithm}
\usepackage{algpseudocode}
\usepackage{xcolor}
\usepackage{tabularx}
\usepackage{multirow}
\usepackage{svg}
\usepackage{tcolorbox}
\usepackage{xcolor}
\usepackage{wrapfig}
\usepackage{pifont}

\definecolor{mygreen}{rgb}{0.0, 0.5, 0.0}
\newcommand{\cmark}{\textcolor{mygreen}{\ding{51}}}
\newcommand{\xmark}{\textcolor{red}{\ding{55}}}

\tcbset{
  promptstyle/.style={
    colback=brown!10,
    colframe=brown!50!black,
    fonttitle=\bfseries,
    coltitle=white,
    colbacktitle=brown!50!black,
    boxrule=0.75mm,
    coltext=black,
    width=\textwidth,
    fontupper=\small
  }
}
\tcbset{
  casestyle/.style={
    colback=purple!10,
    colframe=purple!50!black,
    fonttitle=\bfseries,
    coltitle=white,
    colbacktitle=purple!50!black,
    boxrule=0.75mm,
    coltext=black,
    width=\textwidth,
    fontupper=\small
  }
}

\usepackage{cleveref}

\NewDocumentCommand{\peixuan}
{ mO{} }{\textcolor{purple}{\textsuperscript{\textit{peixuan}}\textsf{\textbf{\small[#1]}}}}

\NewDocumentCommand{\jingjun}
{ mO{} }{\textcolor{blue}{\textsuperscript{\textit{jingjun}}\textsf{\textbf{\small[#1]}}}}

\NewDocumentCommand{\yingjie}
{ mO{} }{\textcolor{orange}{\textsuperscript{\textit{yingjie}}\textsf{\textbf{\small[#1]}}}}

\newcommand\icon{\raisebox{-10pt}{\includegraphics[width=2em]{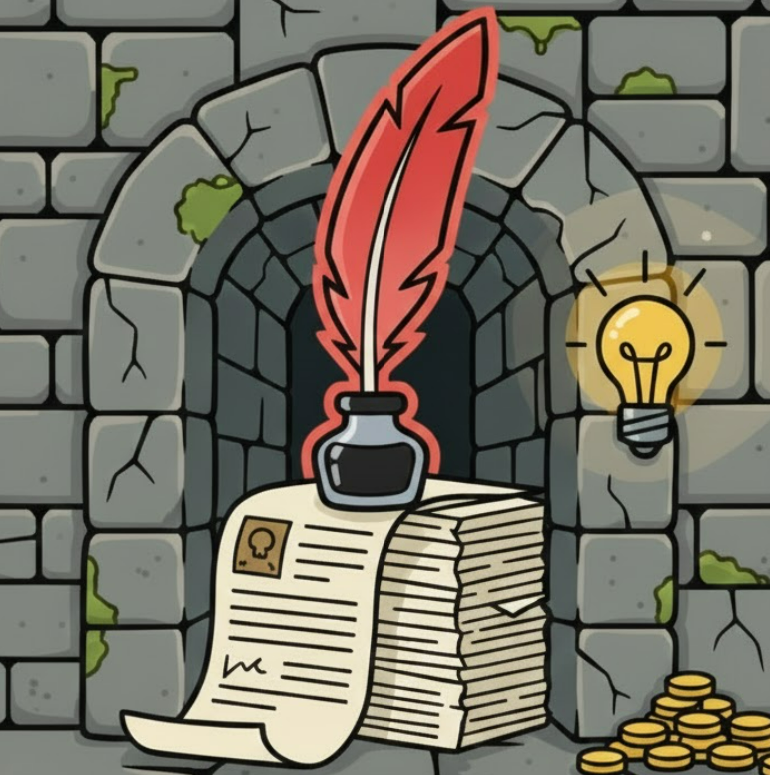}}}
\title{\icon\ DRPG (Decompose, Retrieve, Plan, Generate): An Agentic\\Framework for Academic Rebuttal}

\vspace{-1em}
\author{
Peixuan Han, Yingjie Yu, Jingjun Xu, Jiaxuan You\\
University of Illinois Urbana-Champaign\\
\texttt{\{ph16,yyu69,jingjunx,jiaxuan\}@illinois.edu}\\}

\begin{document}
\maketitle

\input{sections/abstract}
\input{sections/introduction}

\input{sections/related_work}
\input{sections/method}
\input{sections/experiment}
\input{sections/analysis}
\input{sections/conclusion}

\bibliography{custom}
\input{sections/appendix}
\end{document}

%% file: sections/abstract.tex
\begin{abstract}

Despite the growing adoption of large language models (LLMs) in scientific research workflows, automated support for academic rebuttal, a crucial step in academic communication and peer review, remains largely underexplored. Existing approaches typically rely on off-the-shelf LLMs or simple pipelines, which struggle with long-context understanding and often fail to produce targeted and persuasive responses.
In this paper, we propose \textbf{DRPG}, an agentic framework for automatic academic rebuttal generation that operates through four steps: \textbf{D}ecompose reviews into atomic concerns, \textbf{R}etrieve relevant evidence from the paper, \textbf{P}lan rebuttal strategies, and \textbf{G}enerate responses accordingly. Notably, the Planner in DRPG reaches over 98\% accuracy in identifying the most feasible rebuttal direction. Experiments on data from top-tier conferences demonstrate that DRPG significantly outperforms existing rebuttal pipelines and achieves performance beyond the average human level using only an 8B model. Our analysis further demonstrates the effectiveness of the planner design and its value in providing multi-perspective and explainable suggestions. We also showed that DRPG works well in a more complex multi-round setting. These results highlight the effectiveness of DRPG and its potential to provide high-quality rebuttal content and support the scaling of academic discussions. Codes for this work are available at \url{https://github.com/ulab-uiuc/DRPG-RebuttalAgent}.

\end{abstract}

%% file: sections/introduction.tex
\section{Introduction}
\label{sec:intro}


With the rapid advancement of large language models (LLMs), AI agents have become increasingly integrated into the human research workflow. In particular, they have begun to assist researchers across multiple stages of scientific discovery, including idea generation~\citep{lu2024aiscientistfullyautomated}, paper writing~\citep{Aydin2025GenerativeAIA}, and peer review~\citep{chang2025treereview,yu2024researchtown}. Despite these advances, \textbf{AI for academic rebuttal}---the process in which authors and reviewers exchange feedback on a paper---remains largely underexplored.

As a critical stage of the research lifecycle, rebuttal plays an essential role in ensuring fair and objective evaluation of submissions. Moreover, as the research community continues to expand, particularly in fast-growing fields such as computer science, preparing thoughtful rebuttals has become increasingly time-consuming for conscientious authors. For instance, major conferences such as NeurIPS and ICLR received over 25,000 submissions in 2025, creating an urgent need for more efficient mechanisms to facilitate communication between authors and reviewers. Consequently, automated or assistive rebuttal agents can potentially reduce researchers’ workload substantially, allowing them to focus more on innovative research.

Despite its potential benefits, automating academic rebuttal with LLMs is a challenging task. Rebuttal represents a unique adversarial, multi-agent scenario that requires diverse skills, including precise comprehension, persuasive argumentation, and domain-specific expertise. Prior work typically relies on off-the-shelf LLMs or ad-hoc pipelines to generate responses~\citep{kirtani2025revieweval,jin2024agentreview}; however, such approaches often yield suboptimal results for two main reasons. Firstly, academic papers are usually lengthy and information-dense. As revealed by the ``Lost in the Middle'' phenomenon~\citep{liu2024lost}, LLMs struggle to identify and extract the most relevant evidence from long contexts when responding to specific reviewer concerns. Secondly, effective rebuttals require well-structured and convincing arguments tailored to reviewers’ critiques. Since LLMs are not explicitly trained for persuasion, they tend to produce responses that are overly generic, excessively conciliatory or defensive, failing to directly and convincingly address reviewers’ key concerns.

To overcome these limitations, we propose \textbf{DRPG}(Decompose, Retrieve, Plan, Generate), \textbf{a four-stage agentic framework designed to automatically generate high-quality academic rebuttals}. To mitigate long-context challenges, DRPG first employs a Decomposer to break a review into several ``points'' where each point is an atomic concern or confusion that needs to be addressed. For each point, a Retriever then selects the most relevant paragraphs from the paper, reducing the input length by over 75\% while preserving critical evidence needed for rebuttal. To further enhance argument quality, we introduce a Planner that explicitly formulates rebuttal strategies before response generation. Inspired by planning techniques in structured debates~\citep{wang2025strategic,han2025tomap}, the Planner proposes multiple rebuttal perspectives and identifies the most promising one that is best supported by the paper content. Trained with compelling human rebuttal data, the planner can effectively identify the most supported perspective with an accuracy of over 98\%.

Experiments conducted on data from top-tier conferences demonstrate that DRPG can effectively address reviewers’ questions, outperforming existing rebuttal pipelines with around 40 points higher Elo score, which implies consistently higher win rates. In addition, DRPG surpasses average human performance using only an 8B model. Further analyses highlight the successful design of the Planner module, and show that it provides a multi-perspective and explainable signal that substantially improves rebuttal quality. Finally, we showed the impressive performance of DRPG on multi-round discussions and conducted a human study to validate our evaluation metrics. Overall, DRPG represents a promising exploration of integrating LLM agents into the peer-review process, with the potential to reshape how authors and reviewers communicate at scale.

%% file: sections/related_work.tex
\section{Related Work}

\textbf{LLM for academic rebuttal.} Recent advancements in AI have significantly impacted various stages of scientific discovery, including idea generation, experimentation, and paper writing~\cite{luo2025llm4sr,lu2024aiscientistfullyautomated,schmidgall2025agent,yuan2025dolphin}. Among these stages, peer review plays a key role in ensuring the quality and credibility of research papers, and has been receiving increasing attention within the AI research community~\citep{zhuang2025large,liang2024can,wei2025ai}. Existing work in this domain has focused primarily on simulating the review process~\cite{yu2024researchtown,bougie2024generative} and training more effective reviewers~\cite{chang2025treereview,kirtani2025revieweval,jin2024agentreview}. However, the rebuttal phase, which is vital for facilitating communication between authors and reviewers, has received relatively little attention. A few recent studies explore this area by collecting real-world rebuttal datasets~\citep{zhang2025re,kennard2022disapere}, using zero-shot LLMs to generate preliminary rebuttals~\citep{kirtani2025revieweval,jin2024agentreview}, and training rebuttal agents by designing pre-defined templates to address different questions~\citep{purkayastha2023exploring,orbach2019dataset}. Based on these studies, we propose a more systematic and effective rebuttal agent in this work.

\begin{figure*}[t]
    \centering
    \includegraphics[width=\textwidth]{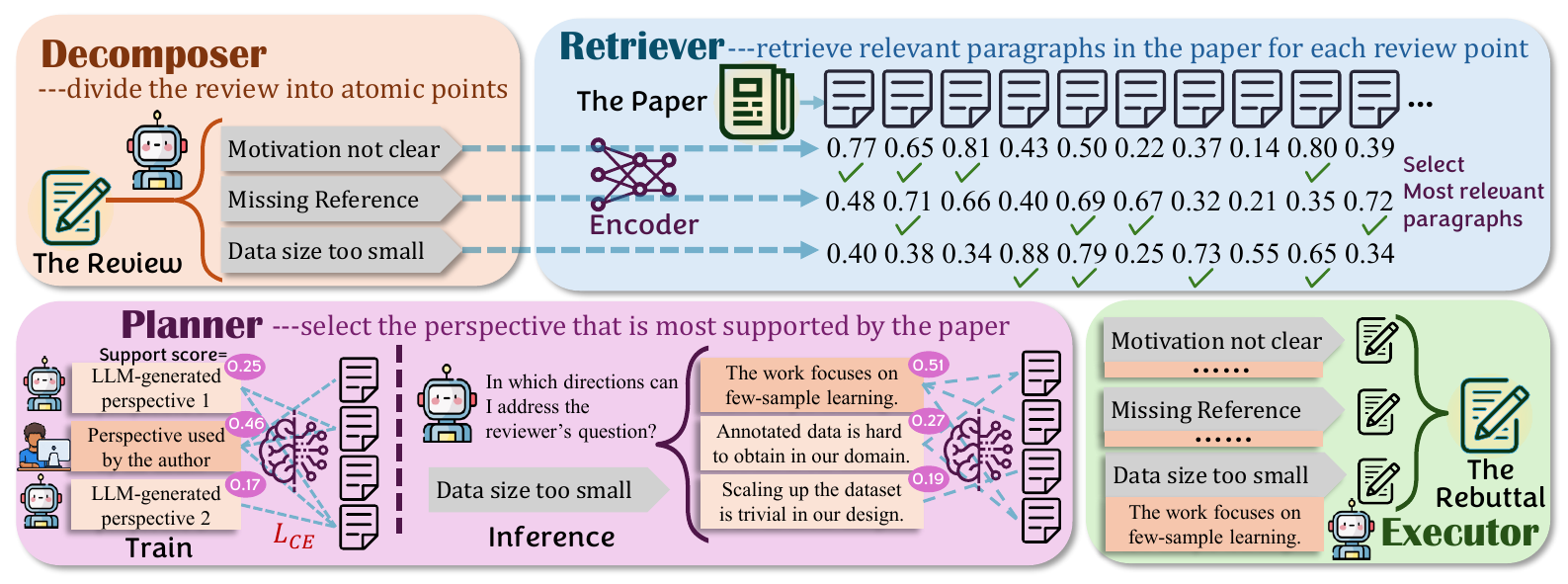}
    \caption{Overview of \textbf{DRPG}.} 
    \label{fig:pipeline}
\end{figure*}

\textbf{LLM for debate and persuasion.} Similar to debate and persuasion~\citep{rogiers2024persuasion}, the objective of the rebuttal is to convince the reviewer to change their opinion. Researchers have utilized human strategies\citep{wang2019persuasion,yang2019let} and high-quality dialogues\citep{singh2024measuring,stengel2024teaching,jin2024persuading,furumai2024zero} to equip LLMs with strong persuasion capabilities. Recently, advanced agentic pipelines~\citep{wang2025strategic,Hu2024DebatetoWriteAPA,Wu2025AIRTA} and Reinforcement learning algorithms \cite{cheng2025towards,han2025tomap} on debate has also been proposed. In the long term, we believe academic rebuttal is a promising domain in debate research, since rebuttal is fact-oriented, grounded in high-quality papers, and has substantial public data available.

\textbf{Planning in high-stakes decision making.} In academic rebuttal, the author must form responses thoughtfully, as the rebuttal quality may affect the result of the submission, with few opportunities to make changes. Planning and selecting argument directions are crucial in such high-stakes settings. There are two prevalent ways of pruning ideas. Firstly, researchers simulate the consequences of adopting each idea, a method originating from Monte Carlo search trees~\citep {coulom2006efficient,silver2016mastering}. This method is most commonly used in scenarios involving multi-agent interactions or external environments~\citep{lu2024llm,shi2024argumentative,weng2024cycleresearcher,he2025debating}. Secondly, researchers train selector or verifier networks to figure out the best candidate through supervised learning~\citep{han2025tomap,wang2025inspiredebate,singh2024enhancing,li2024learning,white2021open,lee2018answerer} or reinforcement learning~\citep{zhang2025echo,chen2025rm} when ground-truth can be obtained at scale. Following this line of work, we train a Planner to select rebuttal perspectives when designing DRPG.





%% file: sections/method.tex
\section{Method}

This section presents an overview of DRPG, an end-to-end framework designed to generate coherent and professional rebuttals based on a full-length, conference-level paper and its reviews. DRPG is composed of four core components—\textbf{Decomposer}, \textbf{Retriever}, \textbf{Planner}, and \textbf{Executor}---each of which is introduced in detail in the following subsections. The overall workflow of the framework is illustrated in \Cref{fig:pipeline}.

\subsection{Decomposer}
Reviews of academic papers typically involve multiple aspects and viewpoints. To transform such multifaceted and complex feedback into manageable points (refer to \Cref{table:perspective_case} for an example) for downstream processing, we employ a \textbf{Decomposer} implemented using a large language model (LLM). The Decomposer identifies the weaknesses and questions raised by the reviewer, which are key elements that must be addressed in the rebuttal. As a result, the Decomposer divides the review into a set of independent, fine-grained points that will be addressed in the following modules.

\subsection{Retriever}
Due to the substantial length of academic papers, the performance of rebuttal agents may be adversely affected by the ``Lost in the Middle'' phenomenon~\citep{liu2024lost}. To address this issue, we divide the paper into paragraphs and employ a \textbf{Retriever} to identify the most relevant paragraphs corresponding to each atomic point generated by the Decomposer. When encoding the paragraphs, we include the hierarchical section headers to capture high-level semantics besides the single paragraph. An example is shown in \Cref{app:case_study}.

The Retriever is implemented using dense retrieval techniques, where a text encoder is used to embed both review points and paper paragraphs, and cosine similarity is applied to measure their relevance. Only the most relevant paragraphs are passed to the executor, thereby increasing information density while reducing content length for around 75\% in practice.

\subsection{Planner}
\label{sec:method:planner}

In academic rebuttal scenarios, identifying an appropriate perspective from which to defend the authors’ work is crucial. However, large language models (LLMs) aren't trained to conduct such deliberate planning, leading them to simply state specific details of the paper while overlooking the reviewer’s underlying reasoning and value judgments. To address this limitation, we introduce a two-step \textbf{Planner} that guides LLMs to explicitly plan how to address review questions, making the communication between authors and reviewers more effective.

\subsubsection{Proposing Rebuttal Perspectives}

In the first step, an \textbf{idea proposer} (implemented as an LLM) generates several candidate perspectives based on each review point. The proposer is instructed to consider two high-level strategies: \textbf{clarification}, which identifies potential misunderstandings in the reviewer’s comments, and \textbf{justification}, which argues that the reviewer’s concern does not invalidate the paper’s core contributions. The two categories are flexible and capture the core assessment of a review --- whether it is technically sound or not\footnote{In practice, ``Admitting Weakness'' and ``Conducting  Experiments'' are also common rebuttal approaches. We opt not to include them for two reasons. Firstly, LLMs tend to admit weaknesses by default. Such a tendency may lead to overly passive responses rather than convincing rebuttals. Secondly, our setting is text-based, so no real experiments could be conducted. “Promising new experiments” could become a shortcut without actually responding to the reviews.}. Note that the paper content is intentionally withheld from the idea proposer to encourage creative and diverse perspective generation. As a result, some proposed perspectives may be infeasible or unsupported, which will be filtered out in the subsequent step.

\subsubsection{Selecting the Valid Perspective}

In the second step, the Planner selects the most suitable perspective by evaluating its \textbf{supportive score} with respect to the paper’s content. Concretely, the Planner is implemented using a text encoder (the same encoder as used in the Retriever) followed by a multi-layer perceptron (MLP). We first obtain vector representations for each candidate rebuttal perspective and each relevant paragraph of the paper using the encoder. These vectors are then concatenated and fed into the MLP to compute a score for each perspective–paragraph pair. The final score of a perspective is obtained by averaging its scores across all relevant paragraphs. Given a perspective ``$\text{pers}$'' and a set of paragraphs $p_{1..K}$ from the paper (where $K$ denotes the number of relevant paragraphs), the supportive score $s(\text{pers}, p)$ is defined as\footnote{The operator $\|$ means vector concatenation.}:
\vspace{-0.5em}

\begin{equation}
\label{eq:planner_score}
\small
s(\text{pers},p)=\frac{1}{K}\sum_{j=1}^K \textbf{M}\big(\textbf{E}(\text{pers}) \| \textbf{E}(p_j)\big),
\end{equation}

where $\textbf{E}$ denotes the text encoder and $\textbf{M}$ represents the MLP module in the Planner.

During training, we select rebuttals that lead to an increase in review scores. For each review point, we construct a candidate set consisting of five ``synthetic'' perspectives generated by the idea proposer and one ``ground-truth'' perspective extracted from the actual content.  The Planner is optimized using a cross-entropy loss. Let the set of candidate perspectives be $I_{1..N}$, and $gt$ denote the index of the ground-truth, the training loss is then defined as:

\begin{equation}
\small
\mathcal{L}(gt) = -\log \frac{\operatorname{exp}\big(s(I_{gt},p)\big)}{\sum_{i=1}^N \operatorname{exp}\big(s(I_i,p)\big)}.
\end{equation}

During inference, we design a self-confidence mechanism to ensure the reliability of the selected perspective. A perspective is passed to the Executor only if its confidence score exceeds a predefined threshold $T$; otherwise, DRPG falls back to the setting without the Planner. The selected perspective and its confidence are computed as:

\begin{equation}
\small
\text{ans}=\operatorname{Argmax}_{i=1}^N s(I_i,p),
\end{equation}

\vspace{-0.5em}
\begin{equation}
\small
\text{conf}(\text{ans})=\frac{\operatorname{exp}\big(s(I_{\text{ans}},p)\big)}{\sum_{i=1}^N \operatorname{exp}\big(s(I_i,p)\big)}.
\end{equation}

\subsection{Executor}
\textbf{The Executor} serves as the final stage of the rebuttal pipeline. Given the structured information produced by the preceding modules, the Executor generates a coherent and persuasive rebuttal paragraph for each individual review point. The Executor can be instantiated using either a general-purpose LLM or a model specialized for rebuttal generation.

In summary, DRPG is an agentic workflow designed to automatically generate high-quality rebuttals. By integrating four specialized components, DRPG addresses two key limitations of using a single LLM for rebuttal writing: the difficulty of effectively processing lengthy paper and review texts, and the tendency to produce generic, insufficiently targeted responses. The overall workflow of DRPG is illustrated in \Cref{alg:1}.

\begin{algorithm}
\small
\caption{The procedure of DRPG\protect\footnotemark.}
\label{alg:1}
\begin{algorithmic}[1]
\Require Paper $P$, Review $R$, Decomposer $\mathbf{D}$, Encoder $\mathbf{E}$, Retrieved count $K$, Idea proposer $\mathbf{I}$, Planner $\mathbf{P}$, Threshold $T$, Executor $\mathbf{X}$

\State $r[1..N_r] \gets \mathbf{D}(R)$ \Comment{DECOMPOSE}

\State $V_p[1..N_p] \gets \mathbf{E}(P[1..N_p])$ \Comment{RETRIEVE}

\State $V_r[1..N_r] \gets \mathbf{E}(r[1..N_r])$

\For{$i = 1$ to $N_r$} \Comment{$p$ means relevant paragraphs}
    \State $sim[i,1..N_p] \gets V_r[i]^\top V_p[j],\ \forall j = 1..N_p$
    \State $p[i,1..K] \gets \text{TopK}(sim[i],K)$
\EndFor
\For{$i = 1$ to $N_r$} \Comment{PLAN}
    \State $I[i,1..N_I] \gets \mathbf{I}(r[i])$ \Comment{candidate ideas}
    \State $s[i,1..N_I] \gets \mathbf{P}(I[i], p[i])$ \Comment{supportive scores (\Cref{eq:planner_score})}
    \State $id \gets \arg\max_j s[i,j])$
    \State $\text{conf} \gets \exp(s[i,id])\ /\ \sum_{j=1}^{N_I}\exp(s[i,j])$
    \If{$\text{conf} \geq T$}
        \State $I_{select}[i] \gets I[i,id]$
    \Else
        \State $I_{select}[i] \gets \epsilon$ \Comment{fallback when conf is low}
    \EndIf
\EndFor
\For{$i = 1$ to $N_r$} \Comment{GENERATE}
    \State $res[i] \gets \mathbf{X}(r[i], p[i], I_{select}[i])$
\EndFor
\State $\text{RES} \gets \big\Vert_{i=1}^{N_r} res[i]$ \Comment{concatenate all responses}
\State \Return $\text{RES}$
\end{algorithmic}
\end{algorithm}

\footnotetext{In \Cref{alg:1}, boldface symbols represent LLMs or networks, and normal symbols represent data variables. Variables starting with $N$ are array sizes induced from LLM outputs.}

%% file: sections/experiment.tex
\definecolor{mypurple}{RGB}{138, 71, 205}

\begin{table*}[htbp]
\caption{Performance of rebuttal agents, which shows \textbf{DRPG generates the most effective rebuttals across all settings}. The last 5 columns in the pairwise comparison section correspond to the results of comparing the different structures of the same base model. Elo scores are calculated within each base model.}
\label{table:result}

\centering
\renewcommand{\arraystretch}{1.0}

\resizebox{\textwidth}{!}{%
\begin{tabular}{l|cccccc|cc}
\toprule
\multirow{2}{*}{Setting} & \multicolumn{6}{c|}{Win Rate Against (\%)}                           & \multirow{2}{*}{\textbf{Elo Score}} & \multirow{2}{*}{\textbf{Judge Score}} \\
                         & \texttt{REAL} & \texttt{Direct} & \texttt{Decomp} & \texttt{DRG} & \texttt{Jiu-Jitsu} & DRPG &                            &                              \\ \hline\hline
\texttt{REAL}                     &   -   &    -       &   -     &         -  &    -   &     -          &          -                  &                      5.72        \\ \hline
\rowcolor{gray!15}
\textbf{Qwen3-8B} (\texttt{Direct})                   &   47.81   &   -        &   40.59     &    37.68       &    42.48   &       34.47        &                       936     &        5.63                      \\
~~~\texttt{Decomp}                &   -   &     59.61     &   -     &      46.77     &   44.11    &       43.37        &                            991&              5.75                \\
~~~\texttt{DRG}                    &   -   &     62.32      &    53.23    &     -      &   45.51   &        41.91       &                         1004   &            5.75                  \\
~~~\texttt{Jiu-Jitsu}                &   -   &     57.52        &   55.89        &   54.49    & - &        41.79       &            1013                &          5.72                    \\
\rowcolor{mypurple!7}
~~~DRPG            &  60.65    &    65.53       &   56.63     &      58.09     &    58.21   &         -      &               \textbf{1054}             &     \textbf{5.78}     \\ \hline
\rowcolor{gray!15}
\textbf{GPT-oss-20B} (\texttt{Direct})                   &   40.28   &   -        &    27.14    &     24.39      &    21.39   &       22.02        &               837             &                     5.73         \\
~~~\texttt{Decomp}                &   -   &      72.86    &   -     &     44.03      &   47.89    &       42.57        &        1012                    &                    5.80          \\
~~~\texttt{DRG}                    &   -   &      75.61     &   55.97     &     -      &   55.53    &     48.02          &                         1054   &                 5.75             \\
~~~\texttt{Jiu-Jitsu}                &   -   &     78.61         &   52.11        &   44.47    &     - &    43.21         &                         1029   &          5.72                    \\
\rowcolor{mypurple!7}
~~~DRPG            &  75.53    &    77.98       &   57.43     &      51.88     &   56.79    &         -      &                           \textbf{1067} &       \textbf{5.88}   \\ \hline
\rowcolor{gray!15}
\textbf{Mixtral-8x7B} (\texttt{Direct})                   &  49.04    &   -        &   17.71     &     12.06      &    12.15   &     13.18          &     738                       &             5.63                \\
~~~\texttt{Decomp}                &   -   &    82.29      &   -     &       31.31    &   28.80    &     27.82          &              959              &               \textbf{5.68}               \\
~~~\texttt{DRG}                    &   -   &     87.94      &  68.69      &     -      &   49.65    &      44.67         &             1088               &           5.66                   \\
~~~\texttt{Jiu-Jitsu}                &   -   &     87.85      &       71.20        &   50.35   & - &      44.41         &        1093                    &                        5.65      \\
\rowcolor{mypurple!7}
~~~DRPG            &  51.42    &      86.62     &   72.18     &       55.33   &     55.59   &         -      &         \textbf{1119}                   &    \textbf{5.68}      \\ \hline
\rowcolor{gray!15}
\textbf{LLaMa3.3-70B} (\texttt{Direct})                   &  50.09    &   -        &   10.10     &      18.27     &    11.49   &      9.07        &     725                       &               5.60               \\
~~~\texttt{Decomp}                &   -   &     89.90     &   -     &      42.78     &   46.74    &     40.61          &                  1040          &               5.67               \\
~~~\texttt{DRG}                    &   -   &     81.73      &   57.22     &     -      &   50.88   &      45.91         &      1065                      &               \textbf{5.68}               \\
~~~\texttt{Jiu-Jitsu}                &   -   &     88.51          &   53.26       &   49.12    & - &       41.28        &          1058                  &                     5.67         \\
\rowcolor{mypurple!7}
~~~DRPG            &   65.44   &      90.93     &   59.39     &      54.09     &   58.72    &         -      & \textbf{1109} & \textbf{5.68} \\ \bottomrule                   
\end{tabular}
}
\end{table*}

\section{Experiments}
\label{sec:exp}


\subsection{Experimental Setup}
\hspace{4pt} \textbf{Dataset.} We conduct experiments on Re$^2$~\citep{zhang2025re}, a large-scale dataset consisting of over 17k academic papers and approximately 60k corresponding reviews and rebuttals collected from 45 top-tier computer science conferences over 8 years, including ACL, ICLR, and NeurIPS. Data statistics are reported in \Cref{app:data}.

\textbf{Models.} We evaluate our method on four base LLMs spanning different families and sizes: Qwen3-8B~\citep{yang2025qwen3}, GPT-oss-20B~\citep{agarwal2025gpt}, Mixtral-8x7B~\citep{jiang2024mixtral}, and LLaMa3.3-70B~\citep{dubey2024llama}. Among all settings, the Retriever is always implemented as BGE-M3~\citep{chen2024bge}.

\textbf{Baselines.} We compare DRPG against both human-written rebuttals from the dataset (denoted as \texttt{REAL}) and four agentic baselines. As summarized in \Cref{table:settings}, the first three baselines, \texttt{Direct}, \texttt{Decomp}, and \texttt{DRG}, correspond to ablated versions of DRPG with specific components removed. In addition, \texttt{Jiu-Jitsu}~\citep{purkayastha2023exploring} serves as a strong baseline that generates perspectives using predefined templates based on question types, replacing the Planner module in our pipeline.

\begin{table}[hbtp]
\caption{Components of different baselines.}
\label{table:settings}
\small
\centering
\renewcommand{\arraystretch}{1.1}
\resizebox{\columnwidth}{!}{%
\begin{tabular}{ccccc}
\toprule
Setting & Decomposer & Retriever & Planner & Executor    \\ \hline
\texttt{Direct} & \xmark & \xmark & \xmark & \cmark \\ 
\texttt{Decomp} & \cmark & \xmark & \xmark & \cmark \\ 
\texttt{DRG} & \cmark & \cmark & \xmark & \cmark \\
\texttt{Jiu-Jitsu} & \cmark & \cmark & \cmark & \cmark \\
DRPG & \cmark & \cmark & \cmark & \cmark \\ \bottomrule
\end{tabular} }
\end{table}

\textbf{Metrics.} We employ two LLM-based evaluation metrics to assess rebuttal quality\footnote{Traditional n-gram-based metrics such as ROUGE and BLEU are not well-suited for evaluating the reasoning quality and coherence of rebuttals, and are therefore omitted.}. First, we use GPT-4o as a pairwise comparator to rank rebuttals and compute an \textbf{Elo score} for each method, following standard practice in open-ended generation tasks~\citep{chiang2024chatbot,boubdir2024elo}. Elo scores are estimated by maximum likelihood under a standard Bradley–Terry model, with a base rating of 1000. To validate the reliability of such comparison, we conduct a human study, the results of which are reported in \Cref{human_study}. Second, we use reinforcement learning to train a judge model based on Qwen3-4B to simulate the reviewer's evaluation and scoring process after reading the rebuttal. The judge model gives \textbf{judge score} exactly identical with human reviewers on 71\% of the test data. Additional details are provided in \Cref{app:judge}.

\textbf{Training and Inference Details.} Detailed training configurations for the Planner are described in \Cref{app:planner}. For retrieval, we set the number of retrieved paragraphs per review point to $K=15$. During inference, the Planner applies a confidence threshold of $T=0.8$. Under this setting, approximately 62\% of review points are assigned a valid perspective. The remaining cases typically fall into two categories. The review point is either straightforward and thus doesn't require an explicit perspective, or it's heavily dependent on specific paper content, making it difficult to propose a valid perspective for the Planner.

\subsection{Main Results}
\label{sec:result}

\Cref{table:result} clearly shows that \textbf{an agentic workflow is crucial for producing high-quality rebuttals}: Directly prompting an LLM to respond to an entire review (\texttt{Direct}) consistently yields inferior performance. All agent-based variants achieve substantially higher scores than \texttt{Direct}, demonstrating the effectiveness of structured processing.

Among all methods, \textbf{DRPG consistently outperforms the other variants} in pairwise comparisons and achieves the highest post-rebuttal scores in most settings. This finding suggests that each component of DRPG plays an important role in mitigating the inherent limitations of a single-LLM approach. Firstly, the Decomposer and Retriever break down complex reviews into atomic, focused points that can be easily handled, avoiding the shortcomings of excessively long contexts. Secondly, the Planner proposes and identifies an appropriate response direction for each review question. Building on these outputs, the Executor can generate high-quality, tailored responses. We show qualitative cases of DEPG's benefit in \Cref{app:case_study}.

Although \texttt{Jiu-Jitsu} adopts a pipeline structure similar to DRPG, its performance is consistently lower due to the limitations of its Planner. Specifically, the Jiu-Jitsu Planner selects from a fixed set of canonical rebuttal templates, which often results in generic or impractical perspectives\footnote{Typical examples include statements such as ``We agree some observations have been made in previous work, but there are critical differences'' or ``We will gladly provide the trained networks on request.''}. In contrast, DRPG employs a content-aware Planner that selects perspectives based on the paper’s content, leading to more specific and persuasive rebuttals.

%% file: sections/analysis.tex
\section{Analysis}

\subsection{Ablation Study on Planner Design}
\label{sec:planner}

The Planner is built around an MLP-based scoring function that selects an effective rebuttal perspective by modeling the \textbf{supportive relationship} between candidate perspectives and paper paragraphs that are relevant to the review point. In this section, we further analyze its design by comparing it with three alternative variants: 1) \texttt{no-paper}, where the MLP scores each perspective independently without taking any paper content as input; 2) \texttt{full-paper}, where all paragraphs of the paper are used instead of the $K$ relevant paragraphs selected by the Retriever; and 3) \texttt{encoder}, a training-free setting which uses vector similarity scores between perspectives and paragraphs as scores.

\begin{table}[H]
\caption{Comparison of different planner designs.}
\label{table:planner}
\small
\centering
\renewcommand{\arraystretch}{1.1}

\begin{tabular}{lcc} \toprule
Setting             & Train loss & Test acc (\%) \\ \hline
\texttt{no-paper}   &     0.5881       &     61.55     \\
\texttt{full-paper} &      0.2393      &     86.41     \\
\texttt{encoder}    &       N/A     &  45.44        \\
Our Planner    &     \textbf{0.0914}       &     \textbf{98.64}    \\\bottomrule
\end{tabular}

\end{table}

\begin{table*}[t]
\caption{Performance of DRPG with restricted perspective types (\textbf{C}larification or \textbf{J}ustification). We use LLaMa3.3-70B as the base model in this experiment.}
\label{table:perspective_analysis}

\centering
\renewcommand{\arraystretch}{1.0}

\resizebox{0.9\textwidth}{!}{%
\begin{tabular}{l|cccc|ccc}
\toprule
\multirow{2}{*}{Setting} & \multicolumn{4}{c|}{Win Rate Against (\%)}                           & \multirow{2}{*}{\textbf{Elo Score}} & \multirow{2}{*}{\textbf{Judge Score}} & \multirow{2}{*}{\textbf{Ratio of Clarification}} \\
                         & \texttt{DRG} & DRPG-C & DRPG-J & DRPG &    &                                  \\ \hline\hline
\texttt{DRG}                       &  - & 49.62  & 64.96  & 45.91  & 1018  & 5.75 &  - \\
DRPG-C                     & 50.38  &   -& 65.95  &  41.49 &  1014 & 5.72 &  100\% \\
DRPG-J                     & 35.04  & 34.05  &  - &  34.02 &  915 & 5.65 &   0\% \\
DRPG                       & 54.09  & 58.51  &  65.98 &  - & \textbf{1051} & \textbf{5.78} & 66.26\%  \\ 
\bottomrule                   
\end{tabular}
}
\end{table*}

\Cref{table:planner} reports the training loss and test accuracy of different planner designs. Our Planner successfully identifies the perspective adopted in successful human rebuttal with an accuracy of \textbf{98.64\%}, substantially outperforming all alternatives. These results lead to the following observations:

\textbullet \hspace{1pt} Incorporating paper content is essential for effective planning. Scoring perspectives in isolation (setting \texttt{no-paper}) results in poor performance.

\textbullet \hspace{1pt} Including the Retriever as a preprocessing step significantly improves performance. Compared to \texttt{full-paper}, using only relevant paragraphs makes the Planner focus on content directly related to the review point, preventing irrelevant paragraphs from dominating the aggregation in \Cref{eq:planner_score}.

\textbullet \hspace{1pt} Simply relying on encoder similarity without learning (setting \texttt{encoder}) is insufficient. This suggests that the relationship between a rebuttal perspective and its supporting evidence is more nuanced than surface-level relevance, and requires a learned module to capture.

\subsection{Analyse on Two Types of Rebuttal Perspectives}

As introduced in \Cref{sec:method:planner}, the Planner considers two types of rebuttal perspectives: \textbf{Clarification} and \textbf{Justification}. Clarification aims to correct factual inaccuracies or misunderstandings in the review, whereas Justification seeks to defend the paper’s methodology or contributions when the reviewer’s comments are factually correct but potentially based on debatable evaluation criteria. \Cref{table:perspective_case} presents an illustrative example of these two perspective types for the same review point.

To analyze the effect of perspective choice, we conduct an ablation study that restricts the Planner to a single perspective type. Specifically, instead of allowing the Planner to select the most supported perspective, we force the Executor to respond using only Clarification or only Justification. We denote these two variants as DRPG-C and DRPG-J, respectively. These settings represent two extreme rebuttal strategies: one that focuses exclusively on factual correctness, and the other that emphasizes significance and contribution.

\begin{table}[h]
\caption{An example of candidate perspectives generated by the Planner.}
\label{table:perspective_case}
\small
\centering
\renewcommand{\arraystretch}{1.1}

\begin{tabular}{p{0.95\linewidth}}
\toprule
\textbf{Example Atomic Point in Real-world Review}: The proposed method performs much worse than HiNet in terms of the extraction accuracy of the secret-in-image hiding. \\
\midrule
\end{tabular}


\begin{tabular}{p{0.63\linewidth} p{0.25\linewidth}}
\toprule
\textbf{Perspective by DRPG} & \textbf{Type} \\ \midrule
PSNR may not be the most suitable metric for evaluating the extraction accuracy in the image hiding task. & Justification \\
Our performance with obfuscating is actually better than HiNet’s. & Clarification \\
Our method works well on secret-in-network hiding, a task much more challenging than image hiding. & Justification \\
\bottomrule
\end{tabular}
\end{table}

As shown in \Cref{table:perspective_analysis}, both variants underperform the full DRPG, and even lag behind the \texttt{DRG} baseline, which does not even include a Planner. This result highlights that relying solely on a single perspective type weakens rebuttal quality. \textbf{Effective academic rebuttals require a balanced use of both clarification and justification.} DRPG adapts between these two strategies depending on the review context: it applies clarification when addressing technical misunderstandings, and justification when responding to critiques based on subjective or questionable evaluation standards.

\subsection{Interpreting Planner Scores}

This section presents an interpretability approach that reveals the explainability advantages of DRPG in the Planner’s decision-making. By examining the Planner's scores for each perspective–paragraph pair individually, we can gain insight into why a particular perspective is selected.

\begin{figure}[htbp]
    \centering
    \includegraphics[width=\columnwidth]{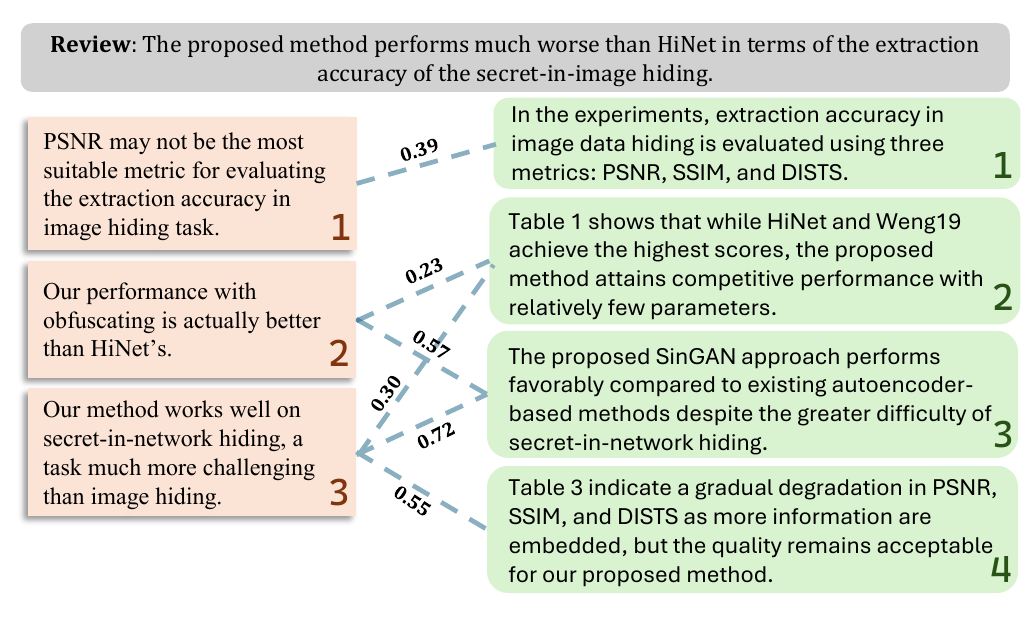}
    \caption{An example to illustrate how the planner evaluates different perspectives. Scores presented are normalized using a sigmoid function, and only scores $\geq 0.2$ are displayed.}
    \label{fig:planner_case}
\end{figure}

\Cref{fig:planner_case} illustrates the Planner’s scores for the example shown in \Cref{table:perspective_case}. Through supervised training, the Planner learns to capture claim–evidence relationships between candidate perspectives and paper content, rather than relying solely on surface-level semantic similarity. For example, perspective 3 is most strongly supported by paragraph 3, which explicitly states that SinGAN performs better on challenging tasks. Paragraph 4 also receives a relatively high score, as it discusses embedding richer information, which can serve as auxiliary evidence when constructing the rebuttal from the angle of ``hard tasks''. Such fine-grained score analysis not only improves the transparency of the Planner’s decision-making process but also provides a useful structural guide for human authors when composing or refining rebuttals.

\subsection{Multi-round Discussion with DRPG}
\label{app:multi_round}

Previous experiments focus on single-round rebuttals, however, in real conference review processes, authors and reviewers sometimes engage in multiple rounds of discussion, during which evaluations of the work become more informed and objective. To better reflect this setting, we design an experiment that simulates multi-round reviewer–author interactions and evaluates the rebuttal agent’s performance accordingly.

The interaction proceeds in a round-by-round manner, alternating between the DRPG and the judge model trained via reinforcement learning (see \Cref{app:judge}). In each round, the judge model first summarizes and evaluates the current rebuttal in its chain-of-thought (CoT), and then outputs a final score. We extract this CoT content as a proxy for the reviewer’s follow-up feedback and treat it as the ``new review'' for the next round. The DRPG then generates a subsequent rebuttal in response. Repeating this process enables us to simulate multi-round discussions between reviewers and authors.

\begin{figure}[htbp]
    \centering
    \includegraphics[width=\columnwidth]{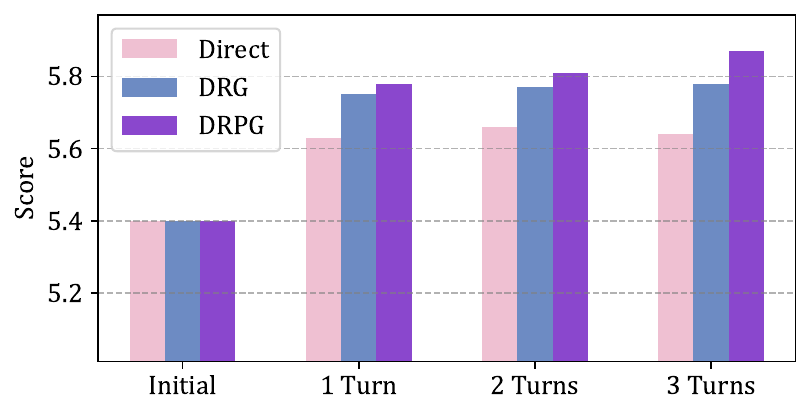}
    \caption{Performance of different rebuttal agents in multi-round discussions. \textbf{DRPG addresses follow-up questions better and delivers greater gains compared with the baselines.}}
    \label{fig:multi_round}
\end{figure}

\Cref{fig:multi_round} illustrates the performance of different workflows over 3 rounds of discussion. As the number of interaction turns increases, the advantage of DRPG becomes increasingly pronounced. While the judge scores of baseline methods quickly plateau after the first round, DRPG continues to achieve consistent improvements in subsequent rounds. This trend suggests that DRPG is better equipped to incorporate feedback from earlier interactions and to respond effectively to follow-up questions raised by the reviewer, enabling reviewers to develop a more complete understanding of the paper’s technical contributions.

\subsection{Human Study on Pairwise Comparison}
\label{human_study}

In this section, we conduct a human study to validate the LLM pairwise comparison in \Cref{sec:exp}. Specifically, we sample 50 reviews and their corresponding rebuttals from \texttt{DRG} and DRPG (the base model is LLaMa-3.3-70B in both settings). The reviews are selected so that exactly 10 \texttt{DRG} rebuttals are preferred by gpt-4o, and 10 DRPG rebuttals are preferred. 3 experts in computer science then independently judge which rebuttal is more effective. 

\begin{table}[hbtp]
\caption{Human study results.}
\label{table:human_study}
\small
\centering
\renewcommand{\arraystretch}{1.1}
\resizebox{\columnwidth}{!}{%
\begin{tabular}{lcc} \toprule
Setting             & Align Rate (\%) & $\kappa$ Score \\ \hline
Among Human Annotators   &    70        & Fleiss's $\kappa=0.598$          \\
Human Majority v.s. LLM &     78       &     Cohen's $\kappa=0.560$    \\ \bottomrule
\end{tabular}
}
\end{table}

In \Cref{table:human_study}, we report the agreement among human annotators, as well as the alignment between human judgments and GPT-4o. The results show that the three human experts exhibit highly consistent preferences, and that their evaluations demonstrate a substantial level of agreement with the LLM. Moreover, qualitative analysis indicates that GPT-4o relies on evaluation criteria similar to those used by human reviewers, further supporting its validity as a high-quality proxy for assessing rebuttal quality. Examples are provided in \Cref{app:case_study}.

%% file: sections/conclusion.tex
\section{Conclusion}

In this work, we investigate the largely overlooked problem of academic rebuttal automation and present \textbf{DRPG}, an agentic framework designed to generate grounded, coherent, and convincing rebuttal responses. DRPG consists of four components: Decomposer, Retriever, Planner, and Executor. By decomposing reviewer feedback, retrieving targeted evidence from long papers, and explicitly planning rebuttal strategies, DRPG addresses key challenges that limit the effectiveness of off-the-shelf LLMs in this setting.
Experimental results on top-tier conference data demonstrate that our approach consistently outperforms existing rebuttal methods and achieves strong performance even with a compact model. Beyond empirical gains, our analysis also shows that structured planning provides an interpretable and multi-perspective signal that meaningfully improves rebuttal quality. As the research community continues to grow, we believe agentic systems like DRPG have the potential to help improve the quality and efficiency of scholarly discussions, thereby supporting the continued development of the academic community.

\section*{Limitation}

This work aims to design an academic rebuttal agent that generates fluent, grounded, and convincing rebuttal arguments. While DRPG shows strong performance, it primarily focuses on clarifying paper content and defending existing contributions, and isn't capable of conducting new experiments. In practice, additional experimental results can sometimes help address reviewers’ concerns during rebuttal. An interesting future direction is to integrate DRPG with AI Scientist systems to support experimental supplementation and achieve complete automation of rebuttal process.

\section*{Ethical Considerations}
This work focuses on automating the academic rebuttal process. While language agents have the potential to significantly assist authors during rebuttal, they also entail inherent risks, particularly those related to hallucinations. Therefore, outputs generated by DRPG (as well as other rebuttal agents) should be carefully reviewed and verified by the authors before submission or release.



%% file: sections/appendix.tex
\onecolumn
\newpage
\twocolumn
\appendix

\section{Prompts}
\label{app:prompt}
This section shows LLM prompts in the paper. 

Prompts used in the DRPG pipeline are shown in \Cref{fig:system_prompt_decomposer,fig:system_prompt_perspective_generator,fig:system_prompt_executor_whole,fig:system_prompt_executor_individual}.
Note that \Cref{fig:system_prompt_executor_whole} is used in the Executor in \texttt{Direct} setting, and \Cref{fig:system_prompt_executor_whole} is used in other settings. After the Executor responds to each individual point, they're merged together to form a complete rebuttal. Refer to \Cref{app:case_study} for illustrative examples.

Prompts used in the evaluation are shown in 
\Cref{fig:system_prompt_rebuttal_judge,fig:system_prompt_compare_rebuttals}.
To avoid the position bias, the order of the two rebuttals is randomly swapped during the comparison.

\section{Data and Training Details}
\label{app:hparams}

\subsection{Data Statistics}
\label{app:data}

\begin{table}[hbtp]
    \small
    \caption{Dataset statistics.}

    \begin{subtable}[t]{\linewidth}
        \centering
        \caption{Size of the dataset.}
        \begin{tabular}{cccc}
            \toprule
            Dataset & Papers & Reviews & Rebuttals \\
            \midrule
            \# Train & 17,814 & 62,211 & 34,024 \\
            \# Eval  & 600    & 2,097  & 1,092  \\
            \bottomrule
        \end{tabular}
        \label{tab:data_statistics}
    \end{subtable}

    \vspace{0.8em} 

    \begin{subtable}[t]{\linewidth}
        \centering
        \caption{Score changes after rebuttal.}
        \begin{tabular}{cccc}
            \toprule
            Dataset &
            \begin{tabular}[c]{@{}c@{}}Score \\ Increased\end{tabular} &
            \begin{tabular}[c]{@{}c@{}}Score \\ Decreased\end{tabular} &
            \begin{tabular}[c]{@{}c@{}}Score \\ Unchanged\end{tabular} \\
            \midrule
            \# Train & 23,642 & 207 & 38,362 \\
            \# Eval  & 745    & 5   & 1,347  \\
            \bottomrule
        \end{tabular}
        \label{tab:predicted_score_statistics}
    \end{subtable}

\end{table}

We summarize the data statistics in \Cref{tab:data_statistics}. 
Note that only a subset of official reviews are responded by the authors.
Each official review includes a score reflecting paper quality, and the scores may be changed during the rebuttal process. However, some initial scores are missing in the dataset. To reconstruct the full rebuttal trajectory, we utilize GPT-oss-120B \citep{agarwal2025gpt} with the prompt in \Cref{fig:prompt_score_recover} to predict initial scores from the rebuttal text and the final scores.

Summary statistics of the rebuttal scores are shown in \Cref{tab:predicted_score_statistics}. The resulting score distributions meet our expectations, and we further validated the predictions through human analysis of several randomly sampled examples.

\subsection{Planner Training}
\label{app:planner}
As described in \Cref{sec:method:planner}, we construct the Planner’s training data by combining five candidate perspectives generated by the idea proposer with one ground-truth perspective extracted from human-authored rebuttals. Using this strategy, we collect 50,000 review points to form the training set and an additional 5,000 review points for evaluation.

The Planner is implemented as an MLP with three hidden layers of sizes 2048, 1024, and 512, respectively. The input layer takes the concatenated vectors of a perspective–paragraph pair with a size of 2048, and the output layer produces a scalar supportive score. Training is conducted for 3 epochs using a batch size of 32 and a learning rate of $5\times10^{-5}$. Due to computational constraints, we freeze the parameters of the BGE-M3 encoder during training and update only the MLP module.

\subsection{Judge Model Training}
\label{app:judge}

\begin{figure}[h]
\small
\vspace{-1em}
\caption{GRPO training configuration for the judge model.}
\label{table:hparams}
\centering
\renewcommand{\arraystretch}{1.1}
\begin{tabular}{lc}
\toprule
\textbf{Hyperparameter}            & \textbf{Value}    \\\hline

Learning rate                & $1 \times 10^{-6}$   \\
Rollout temperature                & $1.0$                \\
KL Coefficient          & $0.001$              \\\hline

Train batch size                   & $64$                \\
Mini batch size                & $32$                 \\
Micro batch size               & $16$                 \\
Training steps                     & $200$                \\
Number of Rollouts                & $4$                  \\
\bottomrule
\end{tabular}
\end{figure}

The judge model is trained with Group Relative Policy Optimization (GRPO)~\cite{shao2024deepseekmath}, and the base model is Qwen3-4B. As shown in \Cref{app:prompt}, the judge model is expected to take careful thinking before generating a final score. The reward is calculated as $r=0.25^{|s-s_g|}$, where $s$ is the judge model's answer, and $s_g$ is the actual score. This means the model will receive a full reward when the predicted score matches exactly with the ground truth, and the reward decreases exponentially as the prediction gap increases. \Cref{table:hparams} shows the hyperparameters during training.

\section{The Jiu-Jitsu Baseline}
\label{app:jiu-jitsu}
We include \texttt{Jiu-Jitsu}~\citep{purkayastha2023exploring} as a planning baseline. Different from our Planner, the Jiu-Jitsu Planner generates perspectives through selecting a canonical rebuttal template.

Given an atomic review point $r_i$, \texttt{Jiu-Jitsu} maps the concern to a canonical rebuttal template via a retrieve-and-rank procedure.
First, it converts $r_i$ into a generated natural-language description $d_i$ of the reviewer concern using a fine-tuned sequence-to-sequence language model, which abstracts away surface wording differences and summarizes the underlying issue.
Second, it assigns $d_i$ to the closest attitude root--theme cluster $z_i$ (where the root corresponds to a reviewing aspect such as \textit{Clarity} and the theme corresponds to a target paper section such as \textit{Experiments}), which is associated with a cluster description $d_{z_i}$.
Third, it uses the rebuttal action label $a_i$ provided in the \texttt{Jiu-Jitsu} resources for each $(d_i, z_i)$ to retrieve the relevant candidate rebuttals for ranking.
Finally, \texttt{Jiu-Jitsu} scores each candidate $r \in R(z_i)$ using the cluster description $d_{z_i}$ and action $a_i$, then selects the top-ranked candidate as the canonical rebuttal, which serves as the rebuttal perspective in our pipeline. \Cref{fig:jiujitsu_example} shows an example of the procedure of \textit{Jiu-Jitsu} baseline and the selected canonical rebuttal.

\section{Case Study}
\label{app:case_study}

\Cref{fig:retrieval} shows an example of the retriever. For a paragraph in Section 4.1, we prepend the titles of Section 4 and Section 4.2 to the paragraph before encoding, enabling high-level understanding of the paper content.

\Cref{fig:comp1,fig:comp2} shows two examples comparing \texttt{Decomp}, \texttt{DRG} and DRPG. We also provide the comments from GPT-4o during the pairwise comparison, from which we can observe the value of the Retriever and Planner in providing more concrete rebuttals.

\begin{figure*}[htbp]
\begin{center}
\begin{tcolorbox}[promptstyle]
You are an experienced researcher in computer science. You have written a conference paper in the field of computer science or AI and received a review. You need to analyse the reviewer's comments. Specifically, identify and list all the weakness points or confusions raised by the reviewer.

\ \ \ \ - You may omit minor issues such as typos, but major comments should all be mentioned.
    
\ \ \ \ - Preferably, extract sentences or words directly from the review. Do not oversimplify the comments.\\
\\
Below is an example of the expected output format:

[

\ \ \ \ "The paper introduced two modules, but lacks ablation study which includes only one of them.",
    
\ \ \ \ "What does the author mean by PPO? Further explain will be helpful.",
    
\ \ \ \ "The experimental results are only shown on 1 newly created environment."
    
]
\end{tcolorbox}
\end{center}
\caption{System Prompt for the Decomposer}
\label{fig:system_prompt_decomposer}
\end{figure*}

\begin{figure*}[htbp]
\begin{center}
\begin{tcolorbox}[promptstyle]
You are an experienced researcher in computer science. You have received a review on a research paper. Your task is to propose up to 5 perspectives to address this point in the rebuttal.

\ \ \ \ - The perspective should either show the reviewer's point wrong, or show that the work is valuable even though the review is correct. Specifically, You MUST consider the following two types of perspectives:

\ \ \ \ \ \ \ \ - Clarification: The reviewer may have factual errors or misunderstood in the paper. For example, they may say something is missing when it's actually present in the paper, or say the methodology is wrong because of a misunderstanding.

\ \ \ \ \ \ \ \ - Justification: Defend your choices and explain why the comment doesn't undermine your paper. For example, they may require an experiment which is unfeasible or unnecessary, or require empirical results for a theoretical paper.

\ \ \ \ - DO NOT propose suggestions or promises for future revision or future work.
    
\ \ \ \ - DO NOT mention specific locations in the paper since you won't be able to access it (e.g. "in section 3.2").\\
\\
Below is an example of the expected output format:

Input: "The paper introduced two modules, but lacks ablation study which includes only one of them."

Output:

[

\ \ \ \ "Clarification: we have actually included such experiment in the paper.",

\ \ \ \ "Clarification: the two modules are dependent on each other and therefore cannot be separated.",

\ \ \ \ "Justification: the ablation study is not necessary as each module has been individually validated in prior work."

]
\end{tcolorbox}
\end{center}
\caption{System Prompt for the Perspective Generator}
\label{fig:system_prompt_perspective_generator}
\end{figure*}

\begin{figure*}[htbp]
\begin{center}
\begin{tcolorbox}[promptstyle]
You are an experienced researcher in computer science. You have written a conference paper in the field of computer science or AI and received a review. You need to write a rebuttal to address the reviewer's comments and convince them to increase their score.\\
\\
Guidelines:

1. Be polite, concise, and professional. Make sure all responses are factual, respectful, and persuasive.

2. Address each comment point-by-point. It's recommended to format the main part of  the rebuttal as: "Question: ...Response: ...". For each point:

3. For each point, you should respond with clear reasoning, and evidence from the original paper, and your professional knowledge.

\ \ \ \ - If the comment has misunderstood the paper or missed some content, clarify the point. If not, defend your choices and explain why this comment doesn't undermine your paper.
   
\ \ \ \ - DO NOT propose suggestions or promises for future revision or future work.

4. Be confident with your paper. Try your best to explain and validate your work, and rebut the concerns raised by the reviewer.

5. Your rebuttal should be concise and no more than 1000 words. You should directly generate a passage without additional comments or thoughts.
\end{tcolorbox}
\end{center}
\caption{System Prompt for the Executor for a Whole Review}
\label{fig:system_prompt_executor_whole}
\end{figure*}

\begin{figure*}[htbp]
\begin{center}
\begin{tcolorbox}[promptstyle]
You are an experienced researcher in computer science. You have written a conference paper in the field of computer science or AI and received a review. You need to write a rebuttal to address the reviewer's comment and convince them to increase their score.\\
\\
Guidelines:

1. Make sure your response is factual, respectful, and persuasive.

2. You should respond with clear reasoning, and evidence from the original paper, and your professional knowledge.

\ \ \ \ - If the comment has misunderstood the paper or missed some content, clarify the point. If not, defend your choices and explain why this comment doesn't undermine your paper.
   
\ \ \ \ - DO NOT propose suggestions or promises for future revision or future work.
   
3. Be confident with your paper. Try your best to explain and validate you work, and rebute the concerns raised by the reviewer.

4. Your rebuttal should be concise and no more than 200 words. You should directly generate a paragraph without additional comments or thoughts.

\end{tcolorbox}
\end{center}
\caption{System Prompt for the Executor for Individual Review Points}
\label{fig:system_prompt_executor_individual}
\end{figure*}

\begin{figure*}[htbp]
\begin{center}
\begin{tcolorbox}[promptstyle]
You are an experienced academic paper reviewer. You will receive a response from the authors addressing your review comments. Your task is to evaluate the response and decide whether to adjust your original score for the paper.

The scoring rubric is from 1 - 10 scale. Certain scores correspond to the following meanings:

\ \ \ \ - 1: The paper has serious flaws, lacks novelty, or is clearly unsuitable for acceptance.

\ \ \ \ - 3: The paper has significant weaknesses or insufficient contributions.

\ \ \ \ - 6: Top 25\% of all submissions. The paper is slightly above the acceptance threshold, with generally solid work, but some limitations.

\ \ \ \ - 8: Top 10\% of all submissions. The paper has a good-quality paper with clear contributions and well-supported results.

\ \ \ \ - 10: Top 5\% of all submissions. The paper makes exceptional contributions and is recommended for spotlight or oral presentation.\\
\\
You should focus on the following criteria when assessing the author's response:

\ \ \ \ - 1. Does the author's response validates their work with clear arguments and coherent logic?
    
\ \ \ \ - 2. Does the author provide sufficient evidence or reasoning to support their claims?

\ \ \ \ - 3. Is the author's response consistent with the content of the original paper?

In addition, please keep in mind that the goal of the response is to CONVINCE the reviewer about the paper, instead of SUGGESTIONS for future work or ADMITTING weakness.

\ \ \ \ - DO NOT consider suggestions, promises, or impacts for future work and revisions when evaluating the responses. Focus on this paper alone.

\ \ \ \ - DO NOT consider tones or emotional appeals, as long as the content is professional. Focus on the logic and reasoning.\\
\\
Then, you should decide whether to change your score based on the author's response.

\ \ \ \ - You should be confident with your original review in most cases. You may increase your score only if the author provides sufficient reasoning that addresses your comments.
    
\ \ \ \ - Do not increase your score based on minor corrections (e.g. typos) or promises on future revisions.
    
\ \ \ \ - If the original score is low, you should be more lenient in increasing the score. If the original score is high, you should hold a higher standard.
    
\ \ \ \ - In most cases, the score change will be small. Large changes, like 2 points, should be rare and well-justified.\\
    \\
As a conclusion, output "My final score is X" where X is your final score (an integer between 1 and 10).
\end{tcolorbox}
\end{center}
\caption{System Prompt for the Rebuttal Judge}
\label{fig:system_prompt_rebuttal_judge}
\end{figure*}

\begin{figure*}[htbp]
\begin{center}
\begin{tcolorbox}[promptstyle]
You are an experienced academic paper reviewer. You will receive a review of an academic paper in computer science, and two responses from the authors.
Your task is to evaluate the responses and decide which response is better.

The response may address the reviewer's several comments. You should compare the responses to each comment individually.
When comparing the responses, you can refer to the following criteria:

\ \ \ \ - 1. Does the author's response validate their work with clear arguments and coherent logic?
    
\ \ \ \ - 2. Does the author provide sufficient evidence or reasoning to support their claims?
    
\ \ \ \ - 3. Is the author's response consistent with the content of the original paper?
    
In addition, please keep in mind that the author isn't allowed to revise the paper afterwards. That is,  the goal of the response is to CONVINCE the reviewer about the paper, instead of SUGGESTIONS for future work or ADMITTING weakness.

\ \ \ \ - DO NOT consider suggestions, promises, or impacts for future work and revisions when evaluating the responses. Focus on this paper alone.

\ \ \ \ - DO NOT consider tones or emotional appeals, as long as the content is professional. Focus on the logic and reasoning.

Please give concrete evidence while being concise. DO NOT repeat or simply summarize the responses' content or similarities; focus on their differences and YOUR ANALYSIS. Output "I think response X (1 or 2) is better" or "I think two responses are similar in quality" at the end of your answer.
\end{tcolorbox}
\end{center}
\caption{System Prompt for Comparing Two Rebuttals}
\label{fig:system_prompt_compare_rebuttals}
\end{figure*}

\begin{figure*}[htbp]
\begin{center}
\begin{tcolorbox}[promptstyle]
\textbf{System}:

You will be given a reviewer–author discussion text and the paper's final score.
Based only on the discussion text and the final score, predict the paper's initial (pre-discussion) review score.

Strictly output a single valid JSON object and nothing else. The JSON must contain only these two fields:

\{

"opinion": "In 2–6 concise sentences, explain your analysis and list the main evidence/signals that support your prediction. If information is insufficient or contradictory, note that uncertainty here",

"initial\_score": "initial score as an integer from 1 to 10"

\}

Hard rules (must follow):

1. **Output only the JSON object** — no extra commentary, no code fences outside the JSON, no explanations. 

2. `initial\_score` must be an integer between 1 and 10.

3. `opinion` must mention 2–4 clear signals or events from the discussion and explain how they affect the score estimate.

4. Do not invent facts outside the provided discussion text. Avoid hallucination.

5. If the discussion is ambiguous or contradictory, state that in `opinion` and then give the most likely integer prediction.

Usually, the reviewer is confident with their review, which means they only raise or decrease scores where there is sufficient evidence. \\

\textbf{User}:

Discussion text: \{discussion\_text \}

Final score: \{final\_score \} / 10
\end{tcolorbox}
\end{center}
\caption{System Prompt to Predict the Initial Review Score}
\label{fig:prompt_score_recover}
\end{figure*}

\begin{figure*}[htbp]
\begin{center}
\begin{tcolorbox}[casestyle]
\textbf{Review point $r_i$:} ``The experimental setup is unclear. Please specify the hyperparameters and training details.''

\textbf{Generated description $d_i$:} ``Missing experimental details and unclear description of training settings.''

\textbf{Predicted root--theme cluster $z_i$:} Root=\textit{Clarity}, Theme=\textit{Experiments}

\textbf{Action label $a_i$:} \texttt{rebuttal\_concede-criticism}

\textbf{Selected canonical rebuttal template $c_i$ (perspective):} ``We apologize for the unclear description of the experimental settings. We will revise the paper to include the missing hyperparameters and training details.''
\end{tcolorbox}
\end{center}
\caption{An example of \texttt{Jiu-Jitsu} procedure to generate rebuttal perspective for a review point.}
\label{fig:jiujitsu_example}
\end{figure*}

\begin{figure*}[htbp]
    \centering
    \includegraphics[width=\textwidth]{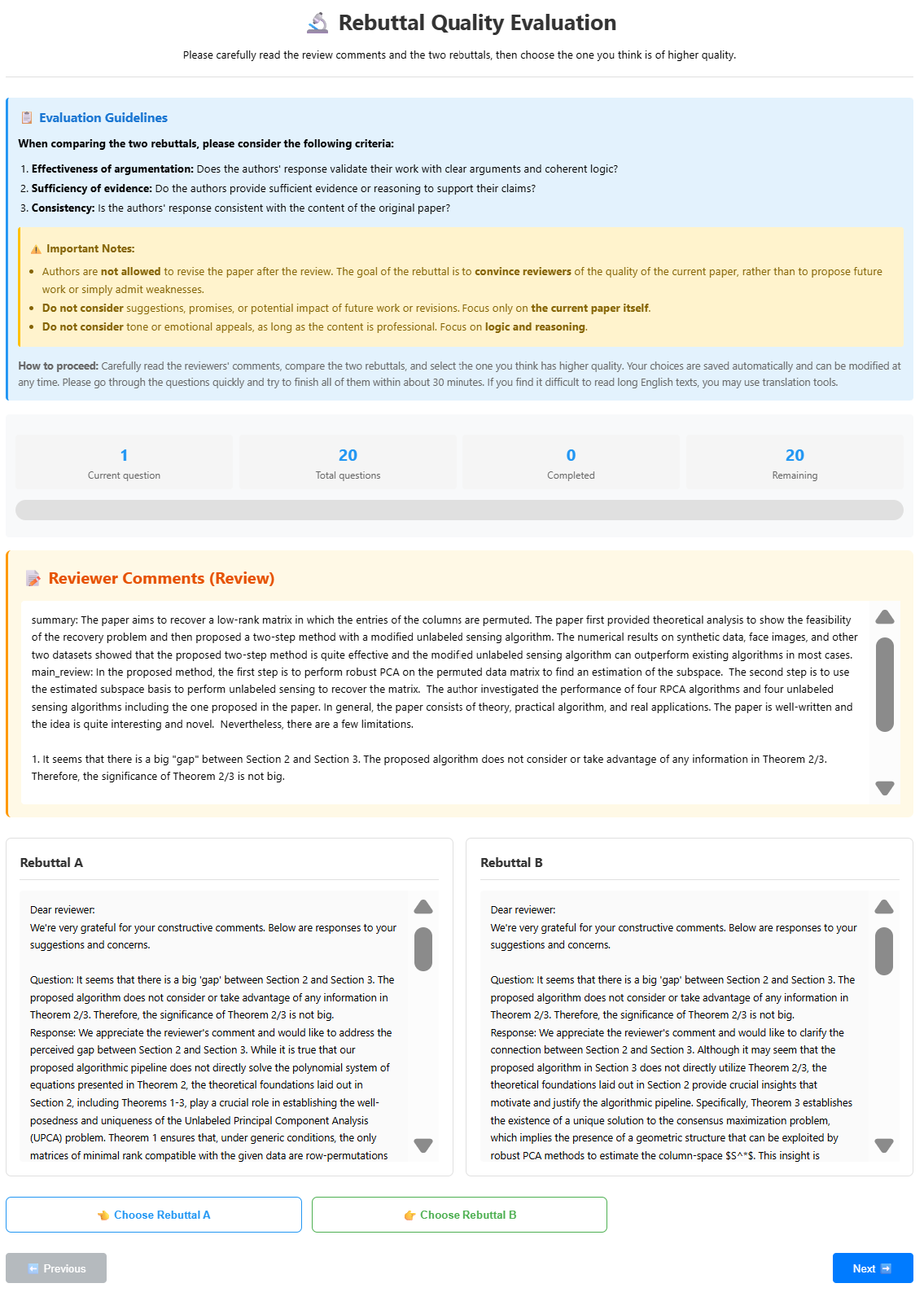}
    \caption{Illustration of the webpage used for human annotation.}
    \label{fig:demo}
\end{figure*}

\begin{figure*}[t]
\begin{center}
\begin{tcolorbox}[casestyle]
\# 4 Experiments

\#\# 4.1 Experimental Setup

Dataset. We conduct experiments on Re$^2$, a large-scale dataset consisting of over 17k academic papers and approximately 60k corresponding reviews and rebuttals collected from 45 top-tier computer science conferences over 8 years, including ACL, ICLR, and NeurIPS. Data statistics are reported in Appendix B.1.
\end{tcolorbox}
\end{center}
\caption{An example of the encoded paragraph format.}
\label{fig:retrieval}
\end{figure*}

\begin{figure*}[t]
\begin{center}
\begin{tcolorbox}[casestyle]
\textbf{Review Point:} Adding factual data in fact may not always lead to clear improvements, while in general most improvement appears to be driven by the heuristic ATE weighting.
          
\textbf{Decomp Response:} We appreciate the reviewer's comment and would like to clarify that the addition of factual data does indeed contribute to the overall improvement, even if its individual contribution may seem marginal in some cases. The comparison between the purple and green bars in Figure 2 and Figure 3, as well as Table 2, shows that the DT policy trained on both factual and counterfactual simulations (purple bar) outperforms the DT policy trained on counterfactual simulations only (green bar). This suggests that the factual data provides some benefit, even if the majority of the improvement comes from the counterfactual simulations and the ATE weighting scheme. Furthermore, the factual data helps to anchor the DT policy in the source environment, making the transfer to the target environment more effective. Therefore, while the ATE weighting scheme is a key component of our approach, the inclusion of factual data is also an important factor in achieving the best results.

\textbf{DRPG Response:} We appreciate the reviewer's observation that the addition of factual data may not always lead to clear visual improvements, as seen in the comparison between the purple and green bars or in Table 2. However, we would like to clarify that the inclusion of factual data is intended to provide a more comprehensive understanding of the results, rather than solely driving improvements. The ATE weighting scheme is indeed a key factor in enhancing the generalization ability of the DT agents, as demonstrated in Figures 2 and 3. The factual data serves as a foundation, allowing the counterfactual simulations to build upon and improve the overall performance. While the improvements may not always be drastic, the combination of factual and counterfactual data with ATE weighting consistently yields better results, as evident in the overall trend of our experimental findings. This suggests that the addition of factual data, although sometimes subtle in its impact, contributes to a more robust and effective training scheme for DT agents.

\textbf{Comment:} Both responses provide coherent arguments and reasoning, but Response 2 has a slight edge due to its emphasis on the foundational nature of the work and its detailed explanation on the role of factual data and ATE weighting.
\end{tcolorbox}
\end{center}
\caption{A case study comparing \texttt{Decomp} amd DRPG.}
\label{fig:comp1}
\end{figure*}

\begin{figure*}[htbp]
\begin{center}
\begin{tcolorbox}[casestyle]
\textbf{Review Point:} The dataset contains only 20 images.

\textbf{DRG Response:} We acknowledge the reviewer's concern regarding the size of the dataset used in our experiments. However, we would like to clarify that the dataset size was intentionally kept small to demonstrate the feasibility and effectiveness of the proposed SinGAN approach in hiding images in deep probabilistic models. With only 20 test image pairs, we were able to achieve promising results in terms of extraction accuracy and model security, as shown in Tables 1 and 2. Furthermore, the use of a small dataset allowed us to focus on the quality and diversity of the generated cover images, as well as the weight distribution similarity between the original and stego SinGANs. While a larger dataset may be desirable for more comprehensive evaluation, our current results provide a solid foundation for future work and demonstrate the potential of the proposed framework.

\textbf{DRPG Response:} We appreciate the reviewer's comment regarding the dataset size. Although our experiments involve a dataset of 20 images, we would like to emphasize that this is sufficient for a proof-of-concept demonstration of our probabilistic image hiding framework. The primary goal of our work is to showcase the feasibility and effectiveness of hiding images in deep probabilistic models, particularly using SinGANs. With a smaller dataset, we are able to establish a baseline for the performance of our approach, as evidenced by the extraction accuracy results in Table 1 and the security evaluation in Table 2. Moreover, our method's ability to hide multiple images for different receivers, as shown in Table 3, further demonstrates its potential. While we acknowledge that larger datasets may be necessary for more comprehensive evaluations, our current results provide a solid foundation for future work, which can indeed focus on scaling up to larger datasets and exploring more complex scenarios.

\textbf{Comment:} Both responses argue that the small dataset size is adequate for a proof-of-concept demonstration. However, Response 2 provides a more detailed explanation by referencing specific tables in the paper that demonstrate the method's effectiveness, thus providing more evidence to support their claim.
\end{tcolorbox}
\end{center}
\caption{A case study comparing \texttt{DRG} amd DRPG.}
\label{fig:comp2}
\end{figure*}